\documentclass[letterpaper]{article} 
\usepackage{booktabs}
\usepackage{aaai2026}  
\usepackage{multirow}
\usepackage{times}  
\usepackage{helvet}  
\usepackage{courier}  
\usepackage[hyphens]{url}  
\usepackage{graphicx} 

\usepackage{natbib}  
\usepackage{subcaption}
 \usepackage{amssymb}
\usepackage{caption} 
\usepackage{amsmath}
\usepackage{capt-of}
\usepackage{algorithm}
\usepackage{algorithmic}

\usepackage{newfloat}
\usepackage{listings}
\DeclareCaptionStyle{ruled}{labelfont=normalfont,labelsep=colon,strut=off} 
\floatstyle{ruled}
\newfloat{listing}{tb}{lst}{}
\floatname{listing}{Listing}
\title{CALYPSO: Forecasting and Analyzing MRSA Infection Patterns with Community and Healthcare Transmission Dynamics }
\author{
    Rituparna Datta\textsuperscript{\rm 1}, Jiaming Cui\textsuperscript{\rm 3}, Gregory R. Madden\textsuperscript{\rm 4}, Anil Vullikanti\textsuperscript{\rm 1,2}
}
\affiliations{
    \textsuperscript{\rm 1}Department of Computer Science, University of Virginia\\
    \textsuperscript{\rm 2}Biocomplexity Institute \& Initiative, University of Virginia\\
    \textsuperscript{\rm 3}Department of Computer Science, Virginia Tech\\
    \textsuperscript{\rm 4}Division of Infectious Diseases \& International Health, University of Virginia Health System\\
}

\newcommand{\anil}[1]{\textcolor{blue}{#1}}
\newcommand{\tool}{CALYPSO}

\newcommand{\toolname}{\textit{\textbf{CAL}ibrated d\textbf{Y}namical \textbf{P}latform for \textbf{S}patiotemporal \textbf{O}utbreaks}}

\begin{document}


\maketitle
\begin{abstract}
Methicillin-resistant \textit{Staphylococcus aureus} (MRSA) is a critical public health threat within hospitals as well as long-term care facilities.
Better understanding of MRSA risks, evaluation of interventions and forecasting MRSA rates are important public health problems.
Existing forecasting models rely on statistical or neural network approaches, which lack epidemiological interpretability, and have limited performance.
Mechanistic epidemic models are difficult to calibrate and limited in incorporating diverse datasets.
We present \tool{}, a hybrid framework that integrates neural networks with mechanistic metapopulation models to capture the spread dynamics of infectious diseases (i.e., MRSA) across healthcare and community settings.
Our model leverages patient-level insurance claims, commuting data, and healthcare transfer patterns to learn region- and time-specific parameters governing MRSA spread. This enables accurate, interpretable forecasts at multiple spatial resolutions (county, healthcare facility, region, state) and supports counterfactual analyses of infection control policies and outbreak risks. We also show that \tool{} improves statewide forecasting performance by over 4.5\% compared to machine learning baselines, while also identifying high-risk regions and cost-effective strategies for allocating infection prevention resources.



\end{abstract}

\section{Introduction}

Hospital-Acquired Infections (HAIs), such as 
Methicillin-resistant \textit{Staphylococcus aureus} (MRSA), pose a heavy health burden, causing over 35,000 deaths annually in the United States~\cite{Prevention.2019}.
This is particularly significant in healthcare facilities, where they contribute to longer hospital stays and increased mortality, leading to increased cost of surveillance, response, control, and treatment, e.g.,~\cite{10.1001/jamainternmed.2013.9763, NEJMoa1801550, weiner:iche2020, 10.1001/jamainternmed.2013.10423}.
Moreover, while often associated with hospitals, HAIs can also spread to non-hospital settings, such as nursing homes, long-term care facilities (LTCFs), and the broader community. LTCFs, in particular, serve as critical reservoirs for MRSA transmission because residents are highly vulnerable and in close contact. 
MRSA colonization (meaning pathogen exposure), has been found to be high in these settings, sometimes $>50\%$ in certain LTCFs and high-risk patient groups~\cite{MRSA_Singapore, MRSA_Transmission_Review, bradley1999methicillin, mody2008epidemiology}. 
Since undetected colonization can drive silent transmission within and between facilities, especially across regions, it is critical to understand and forecast HAI risk, identify high-risk areas, and evaluate the potential impact of infection control policies.
\begin{figure}[t]
    \centering
    \includegraphics[width=1\linewidth]{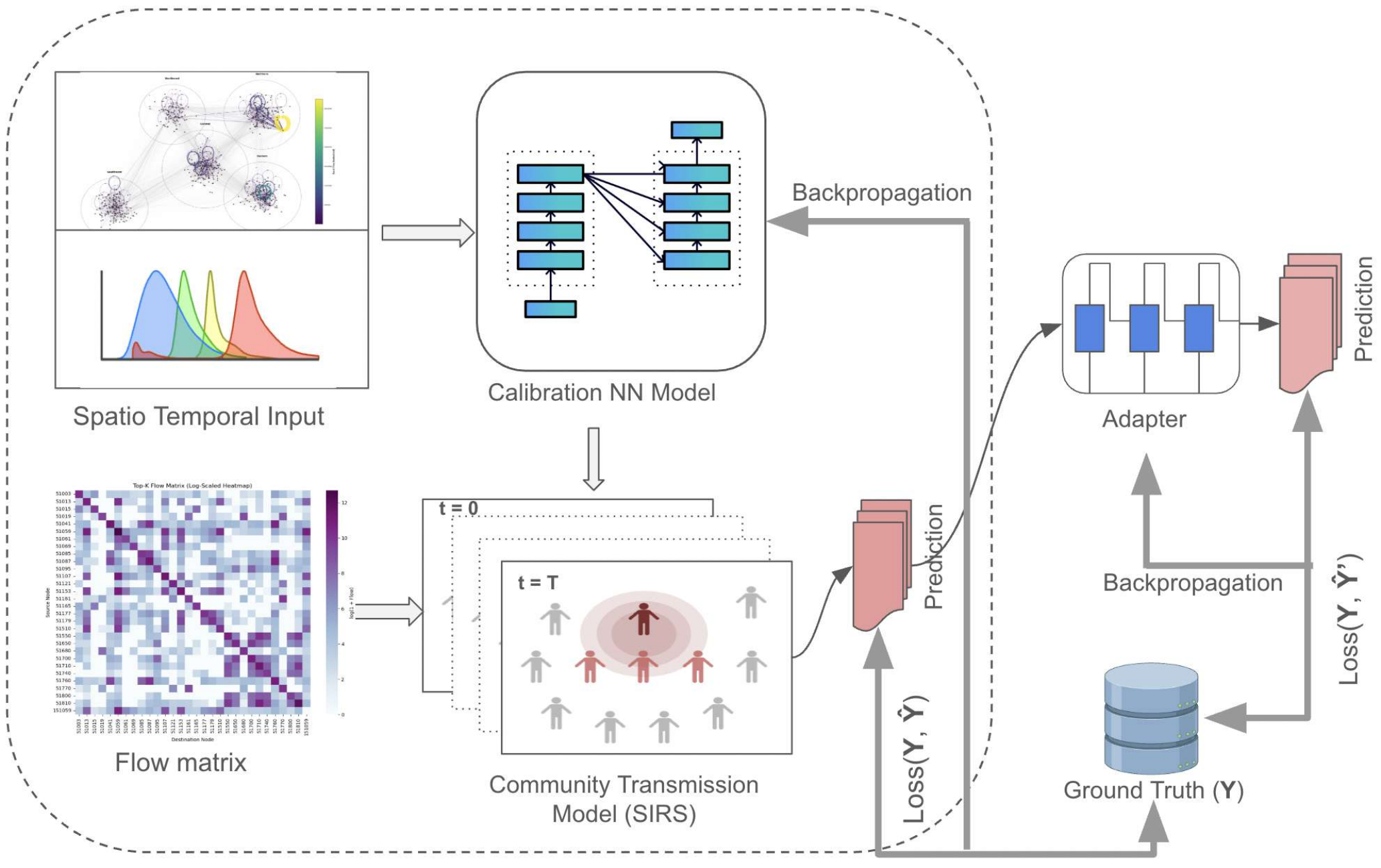}
\caption{Overview of our joint forecasting and dynamical modeling framework, \tool{}. 
A neural network generates region-specific parameters for a metapopulation simulator. 
The adapter refines forecasts at multiple scales.}
    \vspace{-1em}
    \label{fig:joint_model}
\end{figure}

Prior work in this area has been done in two separate directions:
(1) use of mechanistic epidemic models for understanding HAI risk, and  on evaluation and optimization of surveillance and infection, prevention, and control (IPC) measures for HAIs, e.g.,~\cite{lin2025contribution, bartsch2021modeling, lee2021long, lee2016potential, toth2017potential, di2019quantifying}; and
(2) Forecasting MRSA incidence, e.g., ~\cite{Nigo2024, shemetov2025forecasting, ballarin2014forecasting}.

Mechanistic models of different types have been used for understanding CA-MRSA (Community-associated MRSA), including differential equation based and agent based models, e.g.,~\cite{lin2025contribution, macal2014modeling, tosas2016evidence, immergluck2019geographic}.
These studies highlight key drivers of spread outside hospitals~\cite{CA-MRSA, robotham2007meticillin}. 
Infection control policies have also been developed within individual hospitals, e.g.,~\cite{cui2024modeling}; however, such results only consider the cost effectiveness within a single hospital, failing to address the implications of such policies within LTCFs and the community.
Moreover, calibration is a key challenge, and such methods achieve a limited fit with observed infection data, e.g.,~\cite{lin2025contribution}.
Many other factors, e.g., rates of Methicillin-Sensitive Staphylococcus Aureus (MSSA) and antibiotic prescriptions, are known to impact MRSA infection rates \cite{fukunaga2016hospital, andreatos2018impact, kavanagh2019control}.
However, most modeling studies have only used MRSA infection rates for model calibration, and it is not easy to learn from other kinds of datasets. 
 
Diverse kinds of community-level MRSA forecasting methods have been developed, including
statistical time-series approaches \cite{ballarin2014forecasting,vidanage2025application}, machine learning with network features \cite{kamruzzaman2024improving, lin2023machine}, and deep learning models using EHR time-series ~\cite{Nigo2024,shemetov2025forecasting,ML_MRSA,cui2024modeling,jimenez2020feature}.
However, these forecasting models are typically developed at specific geographic resolutions, such as counties or health districts, and often perform poorly when applied across different scales. 
Some related work has focused on forecasting at the individual patient level within hospitals~\cite{Nigo2024, ML_MRSA, cui2025identifying}, which, while valuable, do not address broader population-level patterns across healthcare and community settings.

\paragraph{Contributions.} 
We introduce a new approach, \toolname{} (\tool{}) for understanding the spatio-temporal dynamics of MRSA infection rates and forecasting them within the community and across healthcare facilities of different types.
\emph{\tool{} integrates both tasks of forecasting and learning MRSA transmission dynamics by using a joint deep learning and structured metapopulation based model}, as shown in Figure \ref{fig:joint_model}, for MRSA transmission within counties and healthcare facilities. 
\tool{} consists of two neural network components:
the first learns parameters of the metapopulation model, and second uses the projections from this model to make the forecasts.
The transmission model, $M( \cdot)$, is based on mixing rates 
to model interactions between \textit{patches} (i.e., counties or healthcare facilities), aligning closely with approaches taken in spatially structured epidemic models.
Further, the architecture for \tool{} is very flexible and is easily able to incorporate diverse and heterogeneous datasets, including clinical datasets, as well as information on demographics and staffing at healthcare facilities; these datasets are not all compatible in terms of shape.
Both the neural networks and the transmission model contribute to better performance; dropping either of the components leads to a significant drop in performance (Section \ref{sec:result}).\\
2. We use \tool{} to forecast MRSA infection rates at counties and healthcare facilities of different types (aggregated at a county level) in Virginia, using a large scale All Payers Insurance Claims dataset (APCD) for Virginia, which contains both public and private health insurance claims data. 
We extract data on MRSA and MSSA infection rates and antibiotic prescriptions from APCD.
We consider forecasts at the level of counties and healthcare facilities, as well as health districts and the state, and our results show that \tool{} gives over $4.5\%$ improvement in performance, compared to machine learning baselines.
We observe that a small amount of uncertainty in inputs at patches causes a significant drop in performance.
Further, given a budget $k$, choosing an optimal subset of $k$ patches where reduction in uncertainty maximizes forecast performance is NP-hard.
We show that a greedy choice gives a near-optimal performance, which has important public health utility.\\
3. The metapopulation based MRSA transmission model, $M(\cdot)$, learned from \tool{} has a better fit, compared to baseline calibration methods.
We further find that the model fit is better when all datasets (including MSSA and prescription rates) are used.
We use $M(\cdot)$ for two classes of epidemiological analyses:
(a) \emph{Evaluation and design of infection control policies}: 
such policies are modeled  through reduction in transmission rates in $M(\cdot)$.
First, we consider region level policies, and find that infection control in Eastern Non-General Healthcare facilities in Virginia are most effective in reducing state level MRSA rates.
We also design a greedy strategy for choosing healthcare facilities to implement infection control policies, and find that this reduces the state level outbreak by about a factor of 35, compared to a random baseline;
(b) \emph{Evaluation of outbreak risk}: we find that outbreaks in Northwest General Health Care Facilities led to the highest statewide infection risk.

Thus, \tool{} is a novel approach which is able to incorporate diverse kinds of datasets (not just MRSA infection rates), and simultaneously gives better forecasts at different scales, and provides novel insights about MRSA infection control policies and outbreak risk.

\section{Related Work}
Related work is summarized here; details are in Appendix~\ref{sec:AdditionalRW_app}.  

\noindent
\textbf{Mechanistic Models.} Modeling and forecasting MRSA and other HAIs is challenging due to antibiotic resistance and persistence in healthcare settings. Compartmental epidemic models (SIR/SIRS) have long been used to simulate infection progression \cite{gumel2019mathematical}, capturing susceptible–infected–recovered transitions within facilities. More recent work emphasizes interconnected systems—LTCFs, correctional facilities, and hospitals—where patient transfers and shared healthcare workers drive spread \cite{collins2019systematic}. For example, \cite{MRSA_Singapore} showed that cross-facility dynamics strongly shape outbreak trajectories. Yet, mechanistic models often struggle to capture these interactions, particularly due to simplified assumptions and difficulties in calibrating disease transmission parameters.

\noindent
\textbf{Machine Learning Models.} Deep learning methods such as LSTMs, GRUs, and transformers show promise in forecasting MRSA positivity from electronic health records \cite{ML_MRSA,Nigo2024}. However, they often lack epidemiological interpretability.

\noindent
\textbf{Hybrid Models.} Hybrid approaches combine mechanistic rigor with data-driven flexibility. Physics-Informed Neural Networks (PINNs) embed differential equations into neural networks \cite{raissi2019physics}, while Differentiable Agent-Based Models provide individual-level simulations with learnable parameters \cite{chopra2022differentiable}. Such integration has been applied to influenza, COVID-19, and resistant pathogens. For calibration, ensemble-based filters like the EAKF \cite{anderson2001ensemble} remain widely used for parameter estimation in nonlinear systems.

\noindent
In summary, early MRSA models emphasized isolated settings, but recent work integrates deep learning, mobility-aware dynamics, and hybrid frameworks. \tool{} advances this direction by fusing data-driven forecasting with mechanistic modeling across multiple spatial scales.

\section{Preliminaries}
Let \(\textbf{Y}_{p,t} \in \mathbb{R}\) denote the number of MRSA cases in spatial unit, patch, \(p \in \{1, \ldots, \mathcal{P}\}\) at time \(t \in \{1, \ldots, T\}\), and let \(\textbf{X}_{p,t} \in \mathbb{R}^d\) represent the associated contextual features (e.g., demographics, healthcare access). Here, a patch refers to a county or healthcare facilities located within a county.
We define the historical time series as \(\mathbf{Y}_{1:T} = \{\textbf{Y}_{p,t}\}_{p=1,t=1}^{\mathcal{P},T}\) and \(\mathbf{X}_{1:T} = \{\textbf{X}_{p,t}\}_{p=1,t=1}^{\mathcal{P},T}\). 
We denote the initial infections at $t=0$ as $I_0$, which are also contained in $\mathbf{X}$.
The spatial structure is organized hierarchically: (1) each region \(r \in \mathcal{R}\) is defined as a union of patches, \(r = \bigcup_{p \in \mathcal{P}} p\); and (2) state $\mathcal{S}$ is defined as a union of regions, \(\mathcal{S} = \bigcup_{r \in \mathcal{R}} r\).
\noindent
Incidence at higher spatial levels is computed via additive aggregation:
\[
\textbf{Y}_{r,t} = \sum_{p \in \mathcal{P}} \textbf{Y}_{p,t}, \quad \textbf{Y}_{\mathcal{S},t} = \sum_{r \in \mathcal{R}} \textbf{Y}_{r,t}
\]




\noindent
\textbf{Problem Statements.}
Given historical data \( \{X_{1:T}, Y_{1:T}\} \), the goals are to 
(1) forecast MRSA incidence at a future horizon \( T+h \) across patch, region, and state levels, 
(2) develop a well calibrated MRSA transmission model to evaluate and optimize infection control policies and identify regions with the highest risk, and
(3) when there is uncertainty in the input features, given a parameter $k$, choose a subset $\mathcal{P}'\subset\mathcal{P}$ of patches with $|\mathcal{P}'|\leq k$, so that accurate inputs from $\mathcal{P}'$ maximize the performance of the forecasts.

The forecasting problem is modeled in the following manner.
We define the forecast target as:
\[
\hat{\mathbf{Y}}_{T+h} = \big[ \mathbf{\hat{Y}}_{\mathcal{P},T+h}, \mathbf{\hat{Y}}_{\mathcal{R},T+h} , \mathbf{\hat{Y}}_{\mathcal{S},T+h}\big]
\]
\noindent
We aim to learn a predictive function \( f_\phi \), parameterized by \( \phi \), such that:\\
\centerline{$\hat{\mathbf{Y}}_{T+h} = f_{\phi}(\mathbf{X}_{1:T}, \mathbf{Y}_{1:T})
$}
which will reduce the multi-resolution forecasting loss 
\[
\mathcal{L}_\phi = w_\mathcal{P} \cdot \mathcal{L}_\mathcal{P} + w_\mathcal{R} \cdot \mathcal{L}_\mathcal{R} + w_\mathcal{S} \cdot \mathcal{L}_\mathcal{S},
\]
where the weights \( w_\mathcal{P}, w_\mathcal{R}, w_\mathcal{S} \in \mathbb{R}_{\geq 0} \) allow control over the relative importance of accuracy at each spatial scale, and \(\mathcal{L}_\mathcal{P}, \mathcal{L}_\mathcal{R},\) and \(\mathcal{L}_\mathcal{S}\) denote losses at the patch, region, and state levels, respectively (Equation: \ref{eq:loss}).
Our approach will jointly learn an epidemic transmission model, which allows us to address the epidemiological questions.

\section{\tool{} Framework}

\subsection{Overview of the Framework}
\tool{} consists of three components (Figure~\ref{fig:joint_model}): \textbf{(1)} Epidemic model $M(\cdot)$, \textbf{(2)} Calibration module, $NN_C$, and \textbf{(3)} Adapter module, $NN_A$.
$NN_C$ is a neural network and serves as a calibration model, taking historical covariates and incidence data as input to predict disease transmission parameters $\Theta_{\text{param}}$ for $M(\Theta_{\text{param}},\cdot)$. 
The GRU based adapter module, $NN_A$,  refines the predictions of the $M(\cdot)$, improving fine-grained temporal accuracy while maintaining consistency across spatial scales.


\subsection{Metapopulation Model, $M(.)$} We use a compartmental SIRS metapopulation model, $M(.)$ to simulate the disease progression and transmission. This simulates the infection dynamics across all patches over time. The simulation is governed by the disease model equations, incorporating spatial interactions between patches via a travel matrix $\theta$, normalized with population $\textbf{P}$,
\centerline{$\mathbf{\hat{Y}} = M(\theta, \Theta_{\text{param}}, \textbf{P}, I_0)$}
$\mathbf{\hat{Y}}$ denotes patch-level infection counts, while the simulation incorporates spatial coupling through movement patterns, specifically commuting flows and healthcare facility transfers, encoded in $\theta$, and $\Theta_{param}$ contains the compartmental model’s disease transmission parameters.

\noindent
\textbf{Travel Matrix, $\theta$.} 
        We construct a mobility matrix ($\theta$) to model spatial transmission, where each element \( \theta_{ij} \) represents the proportion of individuals from patch \( i \) traveling to patch \( j \) via county-to-county commuting and healthcare facility transfers. County flows \( C_{ij} \) come from the ACS 2011–2015 \cite{acs}, facility flows \( F_{ij} \) from insurance claims, and patch populations \( \textbf{P}_i \) from census data or claims estimates.
\[
\theta_{ij} = \frac{C_{ij} + F_{ij}}{\textbf{P}_i}, \quad \theta_{ii} = 1 - \sum_{j \ne i} \theta_{ij}
\]  

\noindent
The travel matrix can be either weekly aggregated or static, the latter constructed from fixed travel data representing flows from source patches to destinations. In our approach, we use the static travel matrix, obtained by averaging travel data across all weekly observations.

\noindent    
\textbf{Disease Model for MRSA (SIRS).}
    The disease model for MRSA is formulated using the Susceptible-Infected-Recovered-Susceptible (SIRS) framework \cite{venkatramanan2019optimizing}. 
    This model accounts for the dynamic transitions between different states of infection over time across multiple patches, such as counties and healthcare facilities.
    
    Let \(S_i\), \(I_i\), and \(R_i\) denote the simulated susceptible, infected, and recovered populations from model $M$, in patch \(i\) at time \(t\). Disease dynamics combine local interactions and spatial coupling via the travel matrix \(\theta\), capturing population movement between counties and healthcare facilities. 


The dynamics of the disease are governed by the following equations, 
using mobility-adjusted populations $(N^{\text{eff}}_i)$ and infections $(I^{\text{eff}}_i)$.


\centerline{$N^{\text{eff}}_i = \sum_j \theta_{ji} \textbf{P}_j, \quad I^{\text{eff}}_i = \sum_j \theta_{ji} I_{j,t}$}
\begin{minipage}[t]{0.3\linewidth}
\small
\[
\begin{aligned}
\beta_j^{\text{eff}} &= \frac{I_{\text{eff}}}{N_{\text{eff}}} \cdot \beta \cdot \big[(1-\kappa)(1-\epsilon) + \epsilon\big]; \\
\Delta S_i &= - \sum_{j=1}^K \theta_{ij} \beta_j^{\text{eff}} S_i + \delta R_i;
\end{aligned}
\]
\end{minipage}%
\hfill
\begin{minipage}[t]{0.35\linewidth}
\small
\[
\begin{aligned}
\Delta I_i &= \sum_{j=1}^K \theta_{ij} \beta_j^{\text{eff}} S_i - \gamma I_i; \\
\Delta R_i &= \gamma I_i - \delta R_i;
\end{aligned}
\]
\end{minipage}

\noindent
Here, \( \theta_{ij} \) are the elements of the travel matrix, capturing the interactions between counties and healthcare facilities.  The effective infection rate \( \beta_j^{\text{eff}} \) is computed by considering the effective population and infection rate $(\beta)$ for each county. Other notations: \(\gamma\) (recovery rate), \(\delta\) (reinfection rate), \(\kappa\) (intervention efficacy), and \(\epsilon\) (symptom probability).

These dynamics are simulated over multiple time steps to model the spread and recovery of MRSA across the population, allowing for the forecasting of future infection trends and evaluation of intervention strategies.

\subsection{Calibration Model, $NN_C$}
We calibrate region- and time-specific disease parameters using a neural network that combines temporal sequences and static metadata to produce dynamically varying inputs for epidemiological models. The architecture follows an encoder–decoder design with multi-head attention to capture temporal dependencies and spatial context. The encoder processes multivariate time series and covariates into latent embeddings, which the decoder, conditioned on normalized time indices, refines via attention before mapping to bounded disease parameters \( \Theta_{\text{param}}\) in a range (e.g., infection and recovery rates) that govern disease progression:
\[\Theta_{\text{param}} = \text{NN}_{\text{C}}(\textbf{X}; \psi_C)\] Here, $\psi$ are the learnable weights of the calibration network and
$\Theta_{\text{param}} \in \mathbb{R}^{\mathcal{R} \times T}$ encodes disease parameters of $M(.)$ such as  $\beta, \gamma, \delta, \kappa$ estimated across $\mathcal{R}$ regions and $T$ time steps. These parameters feed into the mechanistic SIRS metapopulation simulator, generating predicted infection dynamics \( \mathbf{Y}' \). 


\subsection{Adapter, $NN_A$}
We introduce an adapter to refine coarse predictions from the broader metapopulation model, improving forecasting precision at state, region, and patch levels. The adapter is a residual GRU module that corrects these fine-scale errors.
\[\hat{\textbf{Y}'} = NN_A( M(.); \psi_A)\] Here, the Adapter-corrected forecasts are denoted by $\hat{\textbf{Y}'}$. It uses normalized timestep embeddings and a multi-layer GRU to capture local dynamics that the primary model misses. The adapter combines \textit{teacher forcing} with \textit{autoregressive prediction} to reduce error, and its modular design enables independent training. Acting as a residual learner, it refines forecasts without retraining the full system.

\subsection{Training}
Training proceeds in two stages:\\
\textbf{1. Joint Calibration.}
The neural network and metapopulation simulator are trained end-to-end by minimizing a weighted mean squared error (MSE) loss across spatial scales:
{\small
\[
\scalebox{0.9}{$ \mathcal{L}(\hat{Y}, Y)= w_\mathcal{P}\mathcal{L}(\mathbf{\hat{Y}}^{(\mathcal{P})}, \mathbf{Y}^{(\mathcal{P})})
+ w_\mathcal{R}\mathcal{L}(\mathbf{\hat{Y}}^{(\mathcal{R})}, \mathbf{Y}^{(\mathcal{R})})
+ w_\mathcal{S}\mathcal{L}(\mathbf{\hat{Y}}^{(\mathcal{S})}, \mathbf{Y}^{(\mathcal{S})})
\label{eq:loss}
$}
\]}

{\small Here, patch loss, \(\mathcal{L}_p = \mathcal{L}(\mathbf{\hat{Y}}^{(\mathcal{P})}, \mathbf{Y}^{(\mathcal{P})})\) (similarly for \(\mathcal{L}_\mathcal{R}, \mathcal{L}_\mathcal{S}\))}


\noindent
1. For each batch, aggregate input features weekly and feed them into the neural network parameter estimator.\\
2. Infer region- and time-specific parameters $(\Theta_{param})$ to drive the metapopulation SIRS simulator.\\
3. Compute the combined MSE loss (Eq.~\ref{eq:loss}) at patch, regional, and state levels.\\
4. Backpropagate loss and update network parameters; save the best model by loss or $R^2$.

Model performance is continuously monitored based on the total loss and the R-squared ($R^2$) score computed on the state-level sum of predictions. The network parameters are saved whenever a new best loss or $R^2$ score is achieved, ensuring that the best-performing model state is preserved throughout training.

\noindent
\textbf{2. Adapter Training.}
With the simulator fixed, adapter is trained to minimize residual forecast errors, improving short-term accuracy.
The full algorithm with notations are detailed in the Appendix (Algorithm \ref{alg:joint_nn_metapop_sirs}, Table \ref{tab:notation}).

\begin{table}
\centering
\caption{Summary of datasets for MRSA forecasting in VA.}
\label{tab:dataset_summary}
\resizebox{\linewidth}{!}{
\begin{tabular}{|p{11cm}|}
\hline
\textbf{Dataset Component and Description. (Details in Appendix \ref{sec:data_app})} \\
\midrule
\hline
\textbf{Insurance Claims (2016--2021):} From the Virginia All-Payer Claims Database (APCD)~\cite{vhi}; 
consists of deidentified patient-level records including:
\begin{itemize} 
    \item \textbf{Identifiers:} Full FIPS county code (mapped to region), visit timestamp.
    \item \textbf{MRSA/MSSA ICD10 Codes:} 
    \texttt{MRSA:} A49.02, B95.62, J15.212, Z22.322, and A41.02. \newline
    \texttt{MSSA}: A49.01, B95.61, J15.211, and Z22.321.
    \item \textbf{Demographics:} One-hot encoded gender (Male/Female), age bands (00–01, 02–04, \dots, 65+), race percentages per infection in historical data.
    \item \textbf{Prescriptions:} Prescription claims per patient listed in the insurance data.
    \item \textbf{Type of Care:} Categorized into \textit{Hospital}, \textit{Outpatient Healthcare Facility}, \textit{Long-term Care}, \textit{General Community}, \textit{Transport}, and \textit{Other}, aggregated from detailed place-of-service codes.
\end{itemize} \\
\midrule
\textbf{Population Data:} County-level population counts from the U.S. Census Bureau, used to normalize infection rates. \\
\midrule
\textbf{Travel Matrix (\(\theta\)):} County-to-county travel volumes within Virginia from American Community Survey commuting data, representing mobility patterns, used to estimate potential transmission across spatial units. \\
\midrule
\textbf{Spatial Mapping:} Virginia is divided into 5 regions (East, Central, Northwest, Southwest, Northern) following Virginia Department of Health classification. Counties are mapped to regions for aggregated modeling~\cite{VDH2025} (Figure \ref{fig:VA_region}, Appendix). \\
\bottomrule
\end{tabular}}
\vspace{-1em}
\end{table}

\section{Results}
\label{sec:result}

We evaluate the performance of \tool{} in three parts:\\
(1) Forecasting accuracy across different spatial resolutions;\\
(2) Value of different components of \tool{} and datasets in MRSA forecasting;\\
(3) Use of the MRSA transmission model for public health analyses, such as evaluating infection control policies and estimating outbreak risk.

\subsection{Experimental Setup}

\noindent
\textbf{Datasets.}
Our primary dataset is the Virginia All Payers Insurance Claims (APCD)~\cite{vhi}, which records patient-level healthcare visits across the state, including timestamps, home counties, and facility attributes. For MRSA-positive patients, demographic covariates are one-hot encoded into fixed-dimensional features. Weekly MRSA case counts are constructed by home county or facility of diagnosis. Since recovery data are absent, each positive case is assumed to persist for a randomly assigned 2–4 weeks, consistent with prior work on MRSA carriage/colonization duration and multisite detection \cite{bradley1999methicillin, mody2003mupirocin}.

Forecasting is conducted at three spatial scales: \textit{statewide}, \textit{region-level}, and \textit{patch-level}. Regions follow Virginia Department of Health divisions, with counts aggregated from constituent counties (Figure~\ref{fig:VA_region}, Appendix). Table~\ref{tab:dataset_summary} details MRSA/MSSA ICD codes and auxiliary demographic and mobility data.





\begin{table}
\setlength{\tabcolsep}{4pt} 
\renewcommand{\arraystretch}{0.9} 
\resizebox{1\linewidth}{!}{%
\begin{tabular}{@{}l c c c c c c@{}}
\hline
Model & Window & Horizon  & R$^2$ & MSE ($\times 10^3$) & MAE & RMSE ($\times 10^3$) \\
\hline
\tool{} & 213 & 4 & 0.92 & 4.69 & 46.54 & 0.069 \\
\tool{} & 261 & 4 & 0.951 & 2.67 & 43.14 & 0.051 \\
\tool{} & 261 & 8 & 0.96 & 2.03 & 34.30 & 0.045 \\
\tool{} & 244 & 4 & 0.95 & 2.11 & 39.38 & 0.046 \\
\tool{} w/o adapter & 213 & 4 & 0.94 & 3.48 & 47.68 & 0.059 \\
\tool{} w/o adapter & 261 & 4 & 0.94 & 3.22 & 45.45 & 0.057 \\
\tool{} w/o adapter & 261 & 8 & 0.95 & 2.80 & 43.56 & 0.053 \\
\tool{} w/o adapter & 244 & 4 & 0.95 & 2.18 & 39.80 & 0.047 \\
LSTM & 213 & 4 & 0.88 & 4.07 & 29.19 & 0.047 \\
LSTM & 261 & 4 & 0.91 & 3.25 & 17.99 & 0.064 \\
$M_{EAKF}(\cdot)$ & 213 & 4 & -0.70 & 110.92 & 265.40 & 0.333 \\
$M_{EAKF}(\cdot)$ & 244 & 4 & -0.30 & 55.24 & 183.48 & 0.235 \\
\hline
\end{tabular}
}
\caption{Model performance metrics comparing simulation to ground truth. MSE and RMSE are reported in $\times 10^3$ units.}
\vspace{-1em}
\label{tab:simulation_metrics}
\end{table}

\noindent
\textbf{Metrics.} We use R-squared $(R^2)$ to assess how well the model explains infection count variability and MSE, MAE, RMSE for error magnitude.

\noindent
\textbf{Baselines.} 
We compare our model’s performance against LSTM, a standard autoregressive neural forecasting model, and a dynamical compartmental model $M_{EAKF}(\cdot)$ calibrated with EAKF \cite{anderson2001ensemble}. 

\noindent
\textbf{Training and Evaluation Setup.}
\tool{} is trained to forecast a 4--8 week horizon with weekly predictions. Input sequences span the full training period, up to $270$ weeks. 
Training runs for 10,000 epochs with Adam (learning rate $5\times10^{-3}$, weight decay 0.01), gradient clipping (10), and StepLR decay (step size 30, $\gamma=0.9$. 
The batch size equals the total number of patches (644). Best checkpoints are chosen by the lowest loss and the highest $R^2$.

\subsection{Forecasting Performance}
\begin{figure}[h!]
    \centering
\includegraphics[width=\linewidth,height=12.5cm]{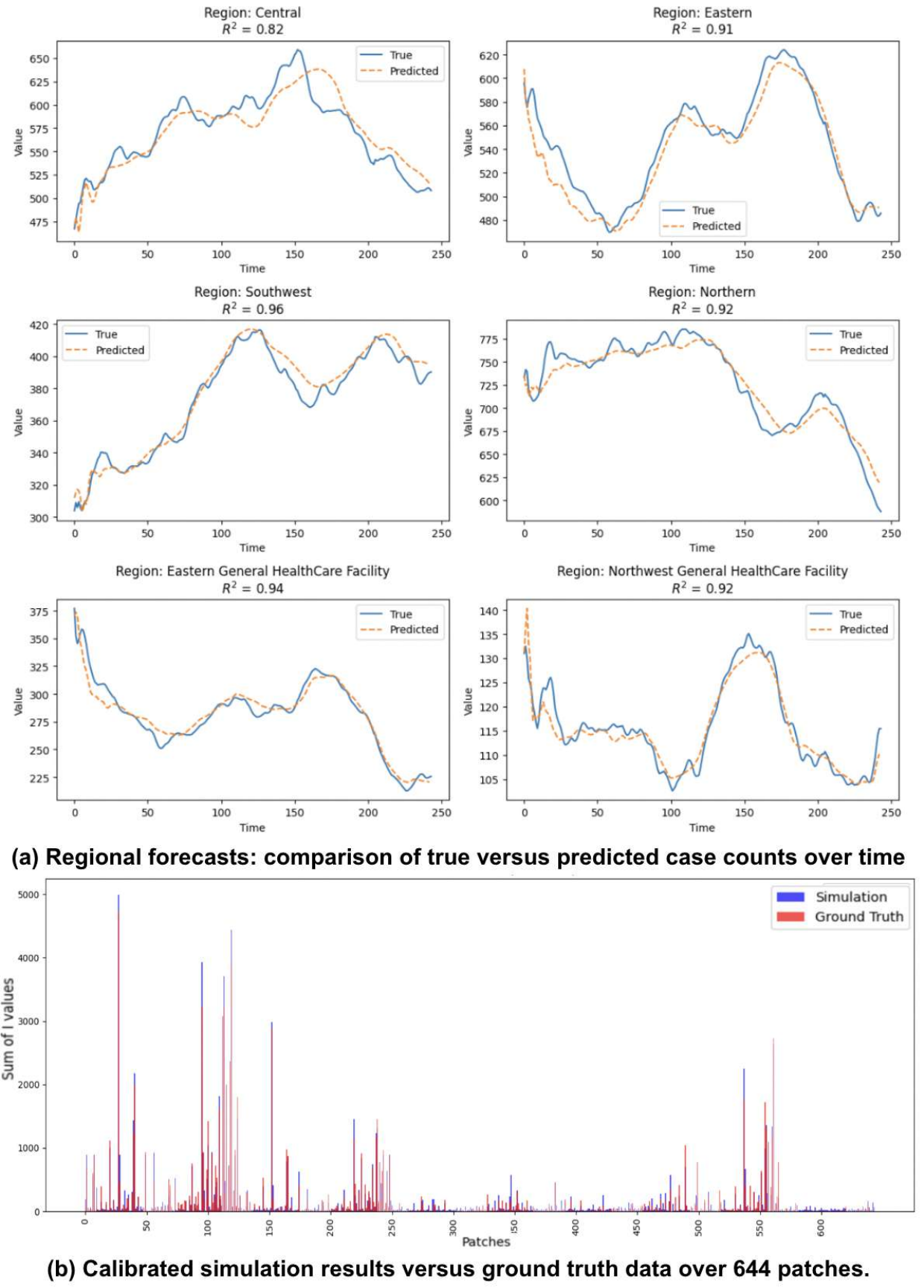}
    \caption{Model calibration and forecasting at the region and patch level}
    \label{fig:region_level}
     \vspace{-1em}
\end{figure}

Our \tool{}'s predictions closely align with the observed ground truth for the summed MRSA counts across all patches over time. Figure \ref{fig:region_level}(a) provides a more granular view of the model's regional forecasting capabilities. The $R^2$ values vary across these forecasts, ranging from  $\approx 0.50$ to $0.96$ (Figure \ref{fig:all_region}, Appendix), demonstrating that while the model achieves high accuracy in many regions, its performance exhibits spatial heterogeneity. To further examine the model's resolution, Figure~\ref{fig:region_level}(b) presents predictions at the individual patch level. Despite increased noise at this finer spatial granularity, the model remains capable of capturing key trends in MRSA case counts. In many patches, the predicted trajectories align well with observed data, demonstrating the model’s ability to generalize across scales. 

\tool{} shows significant improvements over the two baselines, LSTM and $M_{EAKF}(\cdot)$, mentioned earlier.
While the neural networks capture temporal trends, they underperform during sudden changes or with less data.
Similarly, the $M_{EAKF}(\cdot)$ model tracks overall outbreak trends but lacks precision in matching true infection dynamics due to parameter uncertainty and mechanistic limitations (Figure \ref{fig:eakf}, Appendix). As shown in Table~\ref{tab:simulation_metrics}, our model consistently outperforms both baselines across multiple metrics and forecast horizons. For instance, at a 4-week horizon with a 261 week window, our framework improves statewide \(R^2\) by approximately $4.5\%$ and reduces MSE by $18\%$ compared to LSTM. Compared to the EAKF-metapopulation model, it reduces MSE by more than $95\%$, demonstrating the advantage of integrating mechanistic modeling with neural parameter estimation. Additionally, for region-wise prediction, the LSTM model's \(R^2\) ranges from approximately \(-0.85\) to \(0.85\) with an average of $0.45$, approximately $47\%$ lower than our model’s average \(R^2\). For region-wise prediction, \tool{}’s lowest $R^2$ is $0.50$ (vs. LSTM’s $-0.85$), and its average regional $R^2$ is $\approx0.66$; about $47\%$ higher than LSTM’s $0.45$ on average.\\ 
\textbf{Optimally reducing the impact of uncertainty.}
We find that uncertainty in the healthcare facility (HF) patches has a significant impact on performance: $\approx \pm 3\%$ noise in the input features at HF patches reduced statewide accuracy to $R^2=0.62$. 
For a budget $k$, we choose subset $\mathcal{P}'\subset\mathcal{P}$ of HF patches greedily, adding a patch $p$ to $\mathcal{P}'$, which maximizes the $R^2$ metric of the resulting model; the $R^2$ metric is a submodular function of $\mathcal{P}'$~\cite{das2008algorithms}.
We find that removing uncertainty for 6 HF patches led to a $21\%$ gain in $R^2$, closing about $60\%$ of the gap to the clean-data baseline (Figure~\ref{fig:HF_correct}, Appendix).
 


\subsection{Epidemic analyses using learned model, ${\textbf{M}(.)}$}
\label{sec:learning_model}


\noindent
\textbf{The Quality of the Learned Model.} The model’s quality is evident in its consistent performance across different temporal windows and forecasting horizons, as presented in Table~\ref{tab:simulation_metrics}. It captures realistic MRSA transmission dynamics, therefore enabling counterfactual analyses to assess intervention strategies, regional vulnerabilities, and outbreak risks. Below, we highlight three applications.\\ Additional visualizations and extended results for these applications are provided in the Appendix.

\paragraph{Evaluating Infection Control Policies (IPC).}
We evaluate the impact of regional hospital-level MRSA IPC and optimization of IPC resources for a fixed budget.

\noindent
\textbf{(1) \emph{Identifying most effective region}.} 
We first examine how targeted interventions in specific regions influence statewide MRSA spread. 
Following prior intervention studies that explore small reductions in transmission to assess impact, using \tool{},  
we reduce the infection rate $\beta$ by $10\%$ in a single region and analyze infection outcomes state-wide \cite{lee2012simulation, lee2011modeling}. The effect on new infections depends on the mobility-weighted infection force:
\[
\scalebox{0.9}{$
\begin{aligned}
\text{inf\_force}_i(t) &= \sum_j \theta_{ij}\,\frac{I_{\text{eff}, j}(t)}{N_{\text{eff}, j}(t)}\,\beta_j(t), \\
\text{newI}_i(t) &= \text{inf\_force}_i(t)\, S_i(t).
\end{aligned}
$}
\]

Reducing the infection rate ($\beta$) in the Eastern Non-General Health Care Facility region led to the most pronounced statewide reduction in infections (Figure~\ref{fig:beta_reduction_effects}, Appendix). However, mobility-driven spillover effects caused localized increases in cases within the Northwest General Health Care Facilities, underscoring the importance of inter-regional connectivity in sustaining transmission.


\noindent
\textbf{(2) \emph{Optimizing Infection Control Policy}.} 
We aim to minimize total state-level infections by allocating interventions across a limited budget ($B$) of healthcare patches. Let \(\textbf{Z}\in\{0,1\}^\mathcal{P}\) denote the allocation vector, \(M\) our learned model, and \(\mathcal{I}\) the intervention set. The objective is \\
\centerline{\(\displaystyle \min_{\textbf{Z}} \; g(\textbf{Z} \mid M, \mathcal{I})\)}
where \(g(\cdot)\) gives predicted infections under allocation \(\textbf{Z}\). We use a UnitGreedy heuristic that iteratively selects patch \(k^*\) with maximum marginal gain by adding \(e_k\), a unit allocation to patch \(k\),
\centerline{$\Delta_k = g( \textbf{Z} \mid M, \mathcal{I}) - g(\textbf{Z} + e_k \mid M, \mathcal{I})$}

\noindent
UnitGreedy achieves a $\sim\!160\times$ speedup over brute force for selecting 2 patches from 644 candidates. With $B=5$ patches ($\beta_k = 0.9 \beta_k$), it reduces infections by over 6,000 compared to random allocation. Increasing the budget from 1 to 5 patches improves reduction from 2,285 to 6,193 cases (Figure~\ref{fig:budget}, Appendix), a nearly $2.7\times$ gain that demonstrates the budget-dependent benefits of targeted interventions.



\paragraph{Sensitivity to Increased Transmission Risk.}

We assess regional vulnerability by applying a local increase in $\beta$ at patch $p_i$ and measuring the resulting per capita infection rise in region $p_j$, \hspace{1em}
$\text{ImpactRatio}_{j,i} = \frac{\Delta I_j(p_i)}{N_j}.$\\
This experiment highlights that the NW General Healthcare Facility and Eastern General Healthcare Facility are the two most sensitive regions, experiencing the largest impacts from upstream increases in $\beta$ 
(Figure~\ref{fig:sensitivity}, Appendix). Results highlight regions with moderate baseline cases that are highly sensitive to transmission changes elsewhere due to mobility and healthcare transfers. Identifying these vulnerable areas supports targeted surveillance and early intervention, improving coordinated epidemic control.

\paragraph{Impact of Outbreaks.}
Understanding the propagation of localized outbreaks within a connected healthcare network is vital for targeted epidemic control and efficient resource allocation. We analyze (1) which counties \( c_i \in \mathcal{C} \), when seeded with \( K \) additional infections, cause the largest increase in statewide infections \(\Delta I_{\text{state}}(c_i)\), and (2) for a target healthcare patch \( A \in \mathcal{P} \), which external patches \( p_i \in \mathcal{P} \setminus \{A\} \) drive the greatest increase \(\Delta I_A(p_i)\) in infections within \( A \). Using our mobility-coupled SIRS metapopulation model with learned parameters \(\Theta_{param}\) and mobility matrix \(\theta\), we simulate these marginal effects to rank regions by outbreak influence. This identifies high-risk areas and transmission corridors for targeted interventions.

\noindent
With K=50 seeded infections, the model attributes the largest statewide increases to the \textit{Southwest} ($\approx 34.5\%$) and \textit{Northern} ($\approx 25.5\%$) regions, followed by Central ($ 21.9\%$), Eastern ($15.9\%$), and Northwest ($14.9\%$). Due to their larger populations and stronger commuter linkages, outbreaks in the Southwest and Northern regions propagate more widely and amplify statewide impact.

\section{Conclusion}
In this work, we presented \tool{}, a hybrid forecasting framework that combines mechanistic SIRS-based metapopulation models with neural network-based parameter inference. By integrating mobility-informed dynamics with temporal deep learning, \tool{} achieves accurate and interpretable forecasts of MRSA infections across multiple spatial resolutions: state, region, and county levels. Through extensive evaluation, we demonstrated \tool{}'s strong predictive performance, ability to simulate realistic counterfactual scenarios for intervention planning, and robustness to structural ablations. Our results highlight the importance of capturing both spatial interactions and disease-specific dynamics for effective infectious disease modeling, demonstrating \tool{} as a powerful tool for real-time public health forecasting and resource optimization in complex healthcare networks.\\
\textbf{Path to Deployment.}
\tool{} can be deployed by linking APCD claims, census data, and mobility matrices into any state health agency pipeline. After an initial training on a particular state's data, the system can run continuously to update forecasts, simulate counterfactuals, and generate IPC allocation recommendations in near real time. The same setup can be replicated in other states with minimal modification, enabling scalable adoption.

\bibliography{aaai2026}
\clearpage

\renewcommand{\thefigure}{S\arabic{figure}}
\setcounter{figure}{0}

\section{Additional Related Work}
\label{sec:AdditionalRW_app}
\noindent
\textbf{Mechanistic Models:} Understanding and forecasting the spread of Methicillin-resistant \textit{Staphylococcus aureus} (MRSA) and other HAIs have been a long-standing challenge in epidemiological research, particularly due to its resistance to antibiotics and persistence in healthcare environments. Traditional models such as compartmental epidemic models (SIR/SIRS) have been foundational in simulating infection progression \cite{gumel2019mathematical}. These models capture transitions among susceptible, infected, and recovered states and have been used extensively in modeling MRSA spread within individual facilities.

A growing body of research recognizes the importance of modeling MRSA across interconnected systems, including long-term care facilities (LTCFs), correctional facilities, and hospitals, due to frequent patient transfers and shared healthcare workers \cite{collins2019systematic}. Lin et al. \cite{MRSA_Singapore} demonstrated how MRSA dynamics across acute and long-term care facilities can significantly influence outbreak trajectories, highlighting the need for multi-scale spatial modeling. However, these mechanistic models often struggle to fully capture inter-facility MRSA spread, due to their reliance on simplified assumptions and limited integration of facility-specific interactions.

\noindent
\textbf{Machine Learning Models:} Incorporating machine learning and deep learning into infectious disease modeling has been an emerging trend. Models such as LSTMs, GRUs, and transformers have shown promise in time-series forecasting for MRSA positivity using electronic health records \cite{ML_MRSA,Nigo2024}. However, these purely data-driven approaches often lack epidemiological interpretability and perform poorly with sparse datasets.

\noindent
\textbf{Hybrid Models:} To address these shortcomings, hybrid approaches have gained traction. For example, Physics-Informed Neural Networks (PINNs) integrate domain-specific differential equations into neural network training to ensure the learned models respect physical or biological constraints \cite{raissi2019physics}. Similarly, Differentiable Agent-Based Models offer individual-level simulations with end-to-end differentiability \cite{chopra2022differentiable}, enabling parameter learning within complex systems. The idea of integrating mechanistic models with neural networks has also been applied in influenza, COVID-19, and antibiotic-resistant pathogens. In terms of model calibration, Kalman filter-based methods have long been used for parameter estimation in nonlinear systems, such as the Ensemble Adjustment Kalman Filter (EAKF) introduced by Anderson \cite{anderson2001ensemble}, which has informed model calibration strategies.

In summary, while earlier work focused on modeling MRSA in isolated environments, modern approaches increasingly emphasize the integration of deep learning, mobility-aware dynamics, and hybrid frameworks. \tool{} contributes to this evolution by fusing data-driven forecasting with mechanistic epidemiological modeling at multiple spatial resolutions.

\section{Additional Data Details}
\label{sec:data_app}
\begin{table}[h]
\resizebox{1.05\linewidth}{!}{
\begin{tabular}{|p{2cm}| p{10cm}|}
\toprule
\textbf{Parameter} & \textbf{Description} \\
\midrule
\(\beta\) & Infection rate controlling MRSA transmission between susceptible and infected individuals. \\
\(\gamma\) & Recovery rate governing transition from infected to recovered state. \\
\(\delta\) & Reinfection rate dictating the return of recovered individuals to the susceptible state. \\
\(\kappa\) & Behavioral factor modifying susceptibility and infection probabilities based on changes such as hygiene practices. \\
\(\epsilon\) & Symptomatic probability influencing the likelihood of infection detection and treatment. \\
\bottomrule
\end{tabular}
}
\caption{Parameters of the SIRS disease model for MRSA.}
\label{tab:disease_params}
\end{table}

In this section, we present the notation, Table \ref{tab:notation}, that will be used throughout the problem statement and in Algorithm \ref{alg:joint_nn_metapop_sirs}. These definitions serve as a reference to ensure consistency and clarity in the subsequent descriptions.
\begin{figure}[h]
    \centering
    \includegraphics[width=1\linewidth]{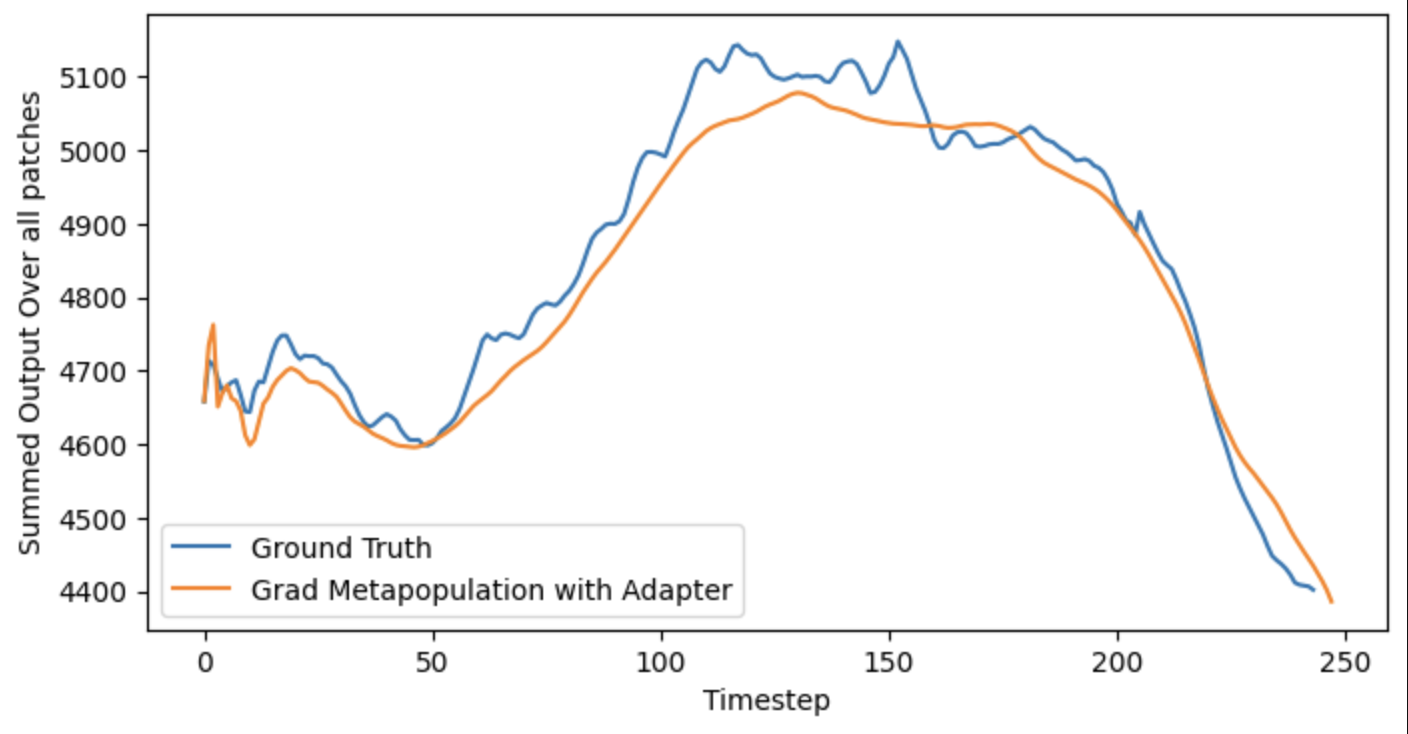}
    \caption{State-Level Fit: Comparison of the overall statewide aggregate MRSA predictions by the \tool{} with Adapter model against ground truth over time.}
    \label{fig:state_level}
\end{figure}
\begin{table}
\centering
\caption{Summary of Notation}
\resizebox{\linewidth}{!}{
\begin{tabular}{ll}
\midrule
\textbf{Symbol} & \textbf{Description} \\
\midrule
$T$ & Historical time steps for training \\
$h$ & Forecast horizon (future time steps) \\
$F$ & Number of input features per patch \\
$\textbf{P}$ & Vector of populations: $P \in \mathbb{R}^N$ \\
$\textbf{X}$ & Input feature tensor: $X \in \mathbb{R}^{N \times T \times F}$ \\
$\textbf{Y}$ & Observed infection counts: $Y \in \mathbb{R}^{N \times T \times 1}$ \\
$\hat{\textbf{Y}}$ & Forecasted infection counts from M(.): $\hat{Y} \in \mathbb{R}^{N \times H \times 1}$ \\
$NN_C$ & Neural network model used for calibration (encoder-decoder) \\
$M(.)$ & Metapopulation Model\\
$NN_A$ & Neural network, Adapter \\
$\theta$ & Static mobility matrix: $\theta \in \mathbb{R}^{N \times N}$ \\
$\mathcal{R}$ & Region-to-patch mapping \\
$\Theta_{params}$ & Disease parameters for all regions $\mathcal{R}$ over time $T$ \\
$\beta_{i,t}$ & Transmission rate for patch $i$ \\
$\gamma_{i,t}$ & Recovery rate for patch $i$ \\
$\delta_t$ & Reinfection rate \\
$\kappa_t$ & Behavioral factor \\
$\epsilon_t$ & Symptomatic rate at time $t$ \\
$I_{i,0}$ & Initial infection seed for patch $i$ \\
$S_{i,t}$ & Number of susceptibles at time $t$ in patch $i$ \\
$I_{i,t}$ & Number of infectious individuals \\
$R_{i,t}$ & Number of recovered individuals \\
$\lambda_i(t)$ & Force of infection for patch $i$ at time $t$ \\
$\Delta I_{i,t}$ & New infections in patch $i$ at time $t$ \\
$N^{\text{eff}}_i$ & Effective population entering patch $i$ \\
$I^{\text{eff}}_i$ & Effective infectious population entering patch $i$ \\
$\mathcal{L_{\phi}}$ & Total loss (multi-resolution) \\
$w_\mathcal{P}, w_\mathcal{R}, w_\mathcal{S}$ & Loss weights for patch/region/state levels \\
\hline
\end{tabular}
}
\label{tab:notation}
\end{table}

\begin{figure}
    \centering
    \includegraphics[width=0.9\linewidth]{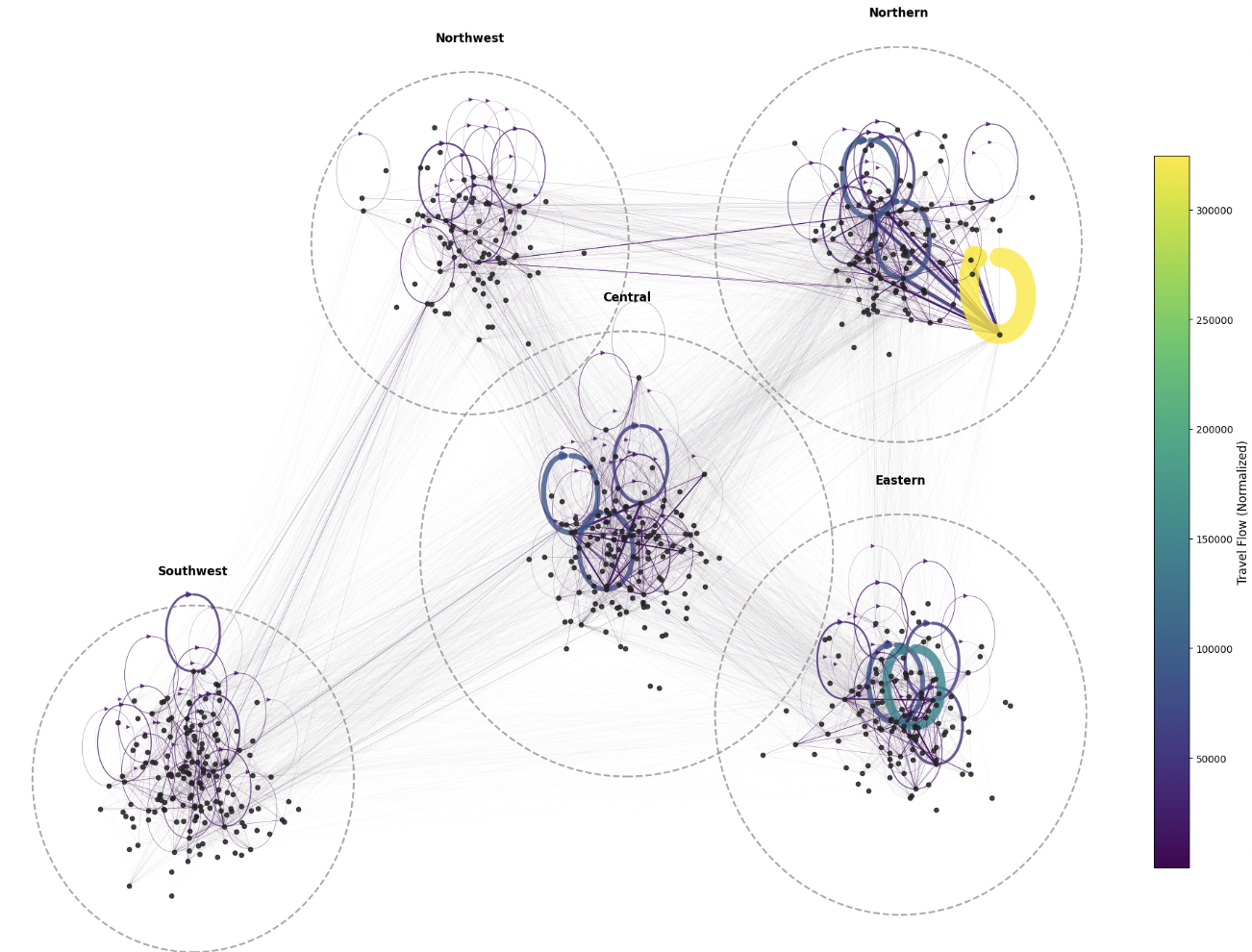}
    \caption{Visualization of the travel Matrix, $\theta$}
    \label{fig:theta}
\end{figure}

\begin{figure}
    \centering
    \includegraphics[width=0.8\linewidth]{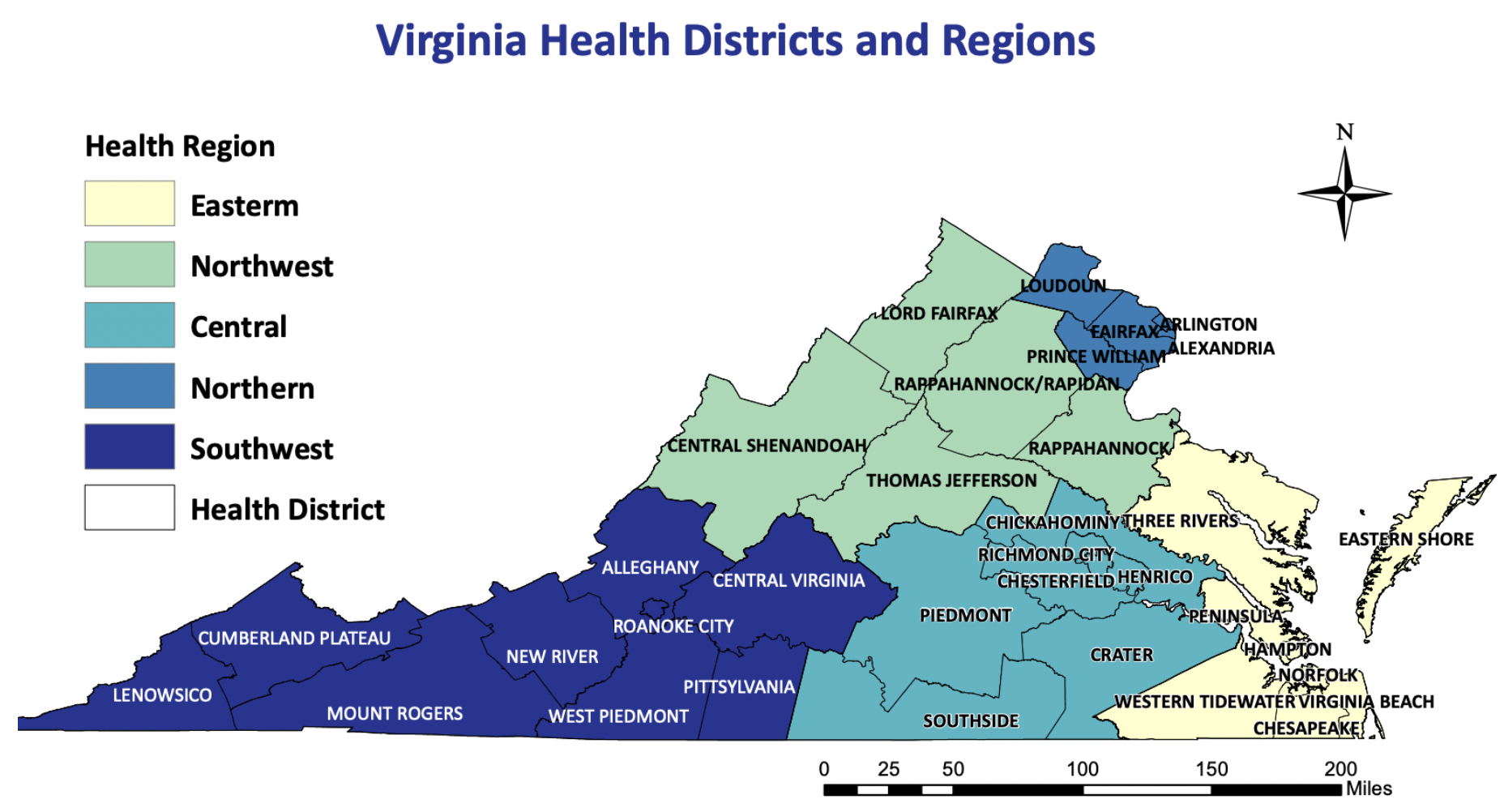}
    \caption{Geographical division of Virginia into five regions—East, Central, Northwest, West, and Northern—based on classifications from the Virginia Department of Health}
    \label{fig:VA_region}
\end{figure}

\paragraph{Patch to Region Mapping}
Figure~\ref{fig:VA_region} shows the Virginia Department of Health (VDH) division of the state into five regions: \textbf{East}, \textbf{Central}, \textbf{Northwest}, \textbf{Southwest}, and \textbf{Northern}. Each patch in our dataset is assigned to one of these regions based on its county location. Beyond geography, each health care patch is also mapped to a broad category:  
\begin{itemize}
    \item \textbf{General Community}
    \item \textbf{Non-General Community} (includes Outpatient Healthcare Facility, Long-Term Care, Hospital, Transport, and Other)
\end{itemize}

Under this scheme, a county location in the general community is labeled as 1\{Region\_Code\} (e.g., Charlottesville as \texttt{NW}). Any hospital in that same county is labeled as 2\{Region\_Code\} (e.g., a Charlottesville hospital as \texttt{2NW}).  

This broad-category labeling collapses fine-grained facility types into a unified \emph{non-general} group while still retaining their geographic region, allowing the model to capture both spatial patterns and the general vs. non-general distinction. We list the healthcare facility types along with their corresponding broad categories in Table~\ref{tab:broad_categories_examples}.

\begin{table}[h]
\centering
\caption{Broad categories with example places of service.}
\renewcommand{\arraystretch}{1.4}
\resizebox{0.52\textwidth}{!}{
\label{tab:broad_categories_examples}
\begin{tabular}{|c|p{3cm}|p{6cm}|}
\hline
\textbf{Category} & \textbf{Broad Category} & \textbf{Examples} \\
\hline
\multirow{1}{*}{1} & General Community & Pharmacy, School, Home, Home-health, Office, Community, Place of Employment, Homeless Shelter,  Birthing center, Temporary Lodging, Mobile Unit \\
\hline
\multirow{5}{*}{2} & Outpatient \newline Healthcare Facility & Walk-in Retail Health Clinic, Urgent Care Facility, Federally Qualified Health Center, Off Campus-Outpatient hospital, Comprehensive outpatient rehabilitation facility, Rural health clinic, Independent Clinic, End stage renal disease treatment facility, Non-residential Opioid Treatment Facility, Mass immunizations center \\
\cline{2-3}
 & Long-Term Care & Assisted Living Facility, Skilled Nursing Facility, Nursing Facility, Hospice,  Community mental health center, Intermediate care facility/mentally retarded, Custodial care facility, Group Home \\
 \cline{2-3}
 & Hospital & Inpatient Hospital, Emergency Room, Psychiatric Facility, Military Treatment Facility,  Independent laboratory, Inpatient psychiatric facility, \\
 \cline{2-3}
 & Transport & Ambulance (Land), Ambulance (Air/Water) \\
 \cline{2-3}
 & Other & Other unlisted facility, Independent laboratory, \\
\hline
\end{tabular}
}
\end{table}

\paragraph{Travel Matrix, $\theta$}
We visualize the patch-to-patch travel flows using a directed network, where nodes represent \textit{patches} and edges indicate \textit{flow} between patches. Edge width and color encode the magnitude of the flow, highlighting high-mobility connections. Nodes are spatially clustered according to their assigned region, with region boundaries drawn as dashed circles for clarity. This layout emphasizes both intra- and inter-regional connectivity patterns.

Figure~\ref{fig:theta} shows this network, illustrating how travel is concentrated within certain regions while still connecting across others. Strong flows are immediately visible due to thicker, darker edges, providing insight into the mobility structure that can influence transmission dynamics.

\begin{algorithm}
\caption{Training and Inference for \tool{}, $f_\phi(.)$}
\label{alg:joint_nn_metapop_sirs}
\small
\begin{algorithmic}[1]
\REQUIRE \begin{itemize}
    \item Historical infection data \( Y_{\text{obs}} \in \mathbb{R}^{N \times T \times 1} \)
    \item Input features \( X_{\text{features}} \in \mathbb{R}^{N \times T \times F} \)
    \item Patch-level population \( P = \{P_i\}_{i=1}^{N} \)
    \item Static connectivity matrix \( \Theta \in \mathbb{R}^{N \times N} \)
    \item Region-to-patch mapping \( \mathcal{R} \)
\end{itemize}
\ENSURE Forecast \( \hat{Y} \in \mathbb{R}^{N \times H \times 1} \)

\vspace{0.5em}
\STATE \textbf{Initialization:}
\STATE Initialize neural network \( f_{\mathrm{NN}} \) to learn disease dynamics
\STATE Initialize metapopulation SIRS simulator \( M_{\mathrm{SIRS}}(P, \Theta) \)
\STATE Define optimizer, learning rate scheduler, loss function \( \mathcal{L} \)

\FOR{epoch \(= 1 \to N_{\text{epochs}} \)}
    \FOR{batch \( (X_b, Y_b) \) from training data up to time \( t \)}
        \STATE Aggregate \( X_b \) to region level using \( \mathcal{R} \)
        \STATE Infer parameters \( \theta^r = f_{\mathrm{NN}}(X_b^r) \)
        \STATE Broadcast \( \theta^r \) to patch-level parameters \( \theta^i \)
        \STATE Simulate: \( \hat{Y}_b = M_{\mathrm{SIRS}}(\theta^i, P, \Theta) \)
        \STATE Compute hierarchical loss:
        \[
        \mathcal{L} = \lambda_1 \cdot \text{MSE}_{\text{patch}} + \lambda_2 \cdot \text{MSE}_{\text{region}} + \lambda_3 \cdot \text{MSE}_{\text{state}}
        \]
        \STATE Update \( f_{\mathrm{NN}} \) via backpropagation
    \ENDFOR
    \STATE Update learning rate
\ENDFOR
\STATE Refine forecasts through $Adapter$

\vspace{0.5em}
\STATE \textbf{Inference:}
\FOR{future timestep \( t+1 \to t+H \)}
    \STATE \( \theta_{t+h} = f_{\mathrm{NN}}(X_{t}) \)
    \STATE \( \hat{Y}_{t+h} = M_{\mathrm{SIRS}}(\theta_{t+h}, P, \Theta) \)
\ENDFOR
\STATE \(\hat{Y'} = Adapter(\hat{\textbf{Y}})\)

\STATE \textbf{Return:} Forecast \( \hat{Y'} \)
\end{algorithmic}
\end{algorithm}

\begin{figure}[h]
    \centering
    \includegraphics[width=0.9\linewidth]{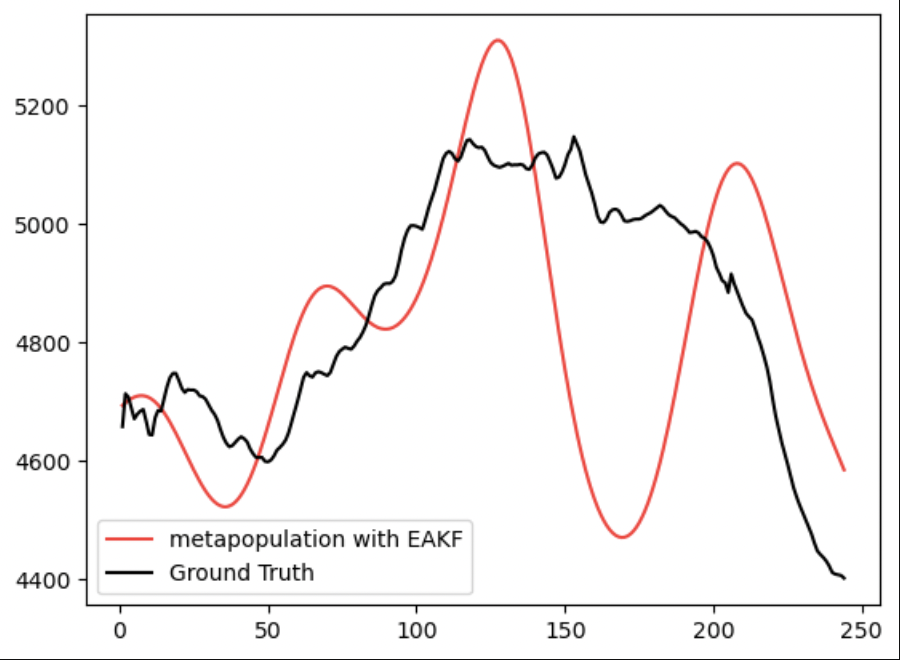}
    \caption{Comparison between EAKF ensemble mean and ground truth infection counts over time.}
    \label{fig:eakf}
\end{figure}

\begin{figure}
    \centering
    \includegraphics[width=0.8\linewidth]{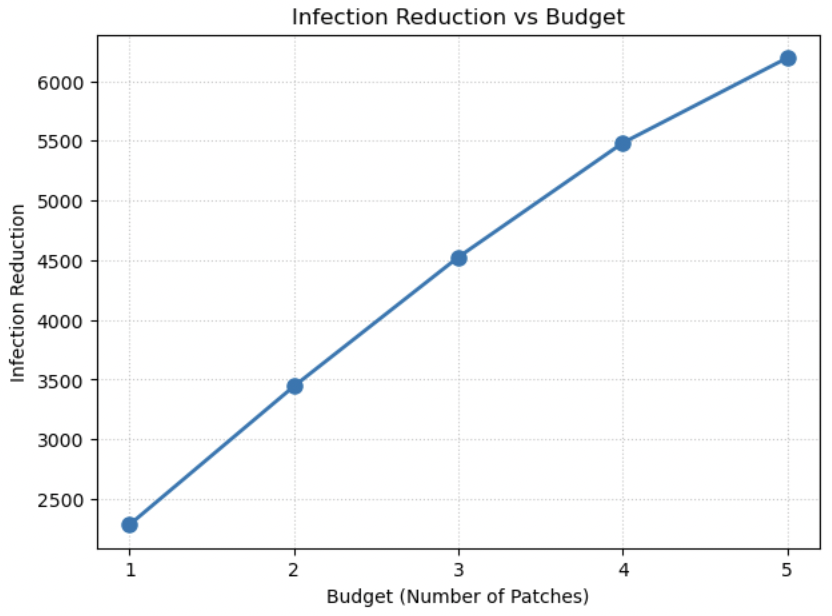}
    \caption{Infection reduction as a function of budget (number of patches selected) with UnitGreedy algorithm}
    \label{fig:budget}
\end{figure}

\begin{algorithm}
\caption{Metapopulation SIRS Simulation,  \(M(\theta^i, P, \Theta) \)}
\label{alg:metapop_sirs_simulator}
\small
\begin{algorithmic}[1]
\REQUIRE 
\begin{itemize}
    \item Patch population \( P \in \mathbb{R}^{N} \)
    \item Parameters: \( \Theta = \left\{\beta_{i,t}, \gamma_{i,t}, \delta_t, \kappa_t, p^{\text{sym}}_t, s_{i,t}\right\} \)
    \item Time horizon \( T \), connectivity matrix \( \theta \in \mathbb{R}^{N \times N} \)
\end{itemize}
\ENSURE Infections \( Y \in \mathbb{R}^{N \times T \times 1} \)

\STATE Initialize: \( S_0 = P - s_0 \), \( I_0 = s_0 \), \( R_0 = 0 \)
\FOR{each time \( t = 0 \to T-1 \)}
    \STATE Compute mobility-adjusted populations:
    \[
    N^{\text{eff}}_i = \sum_j \theta_{ji} P_j, \quad I^{\text{eff}}_i = \sum_j \theta_{ji} I_{j,t}
    \]
    \STATE Compute infection force:
    \[
    \lambda_i(t) = \sum_j \theta_{ij} \cdot \beta_{j,t} \cdot \frac{I^{\text{eff}}_j}{N^{\text{eff}}_j} \cdot \left( (1 - \kappa_t)(1 - p^{\text{sym}}_t) + p^{\text{sym}}_t \right)
    \]
    \STATE New infections:
    \[
    \Delta I_{i,t} = \min(S_{i,t}, \lambda_i(t) \cdot S_{i,t})
    \]
    \STATE Update compartments:
    \[
    \begin{aligned}
    S_{i,t+1} &= S_{i,t} - \Delta I_{i,t} + \delta_t \cdot R_{i,t} \\
    I_{i,t+1} &= \Delta I_{i,t} + (1 - \gamma_{i,t}) \cdot I_{i,t} \\
    R_{i,t+1} &= \gamma_{i,t} \cdot I_{i,t} + (1 - \delta_t) \cdot R_{i,t}
    \end{aligned}
    \]
\ENDFOR
\STATE \textbf{Return:} Infection history $\{I_{i,t}\}$ for all patches $i$ and times $t$
\end{algorithmic}
\end{algorithm}

\section{Additional Result}
\noindent
\textbf{Statewide fit}
We perform forecasting on statewide infection counts, and Figure \ref{fig:state_level} shows the overall fit against the ground truth. In this setting, we achieve an $R^2$ of approximately 0.94, indicating a strong agreement between the predictions and the observed data.

\subsection{EAKF Baseline Performance} 
Figure \ref{fig:eakf} presents results from the EAKF-based mechanistic baseline compared to observed infection counts. While the ensemble mean tracks broad outbreak trends, it shows clear deviations from ground truth due to parameter uncertainty and calibration challenges. The contrast highlights \tool{}’s improvement in resolving infection dynamics through neural network–guided calibration.

\begin{figure}
\centering
\includegraphics[width=1\linewidth]{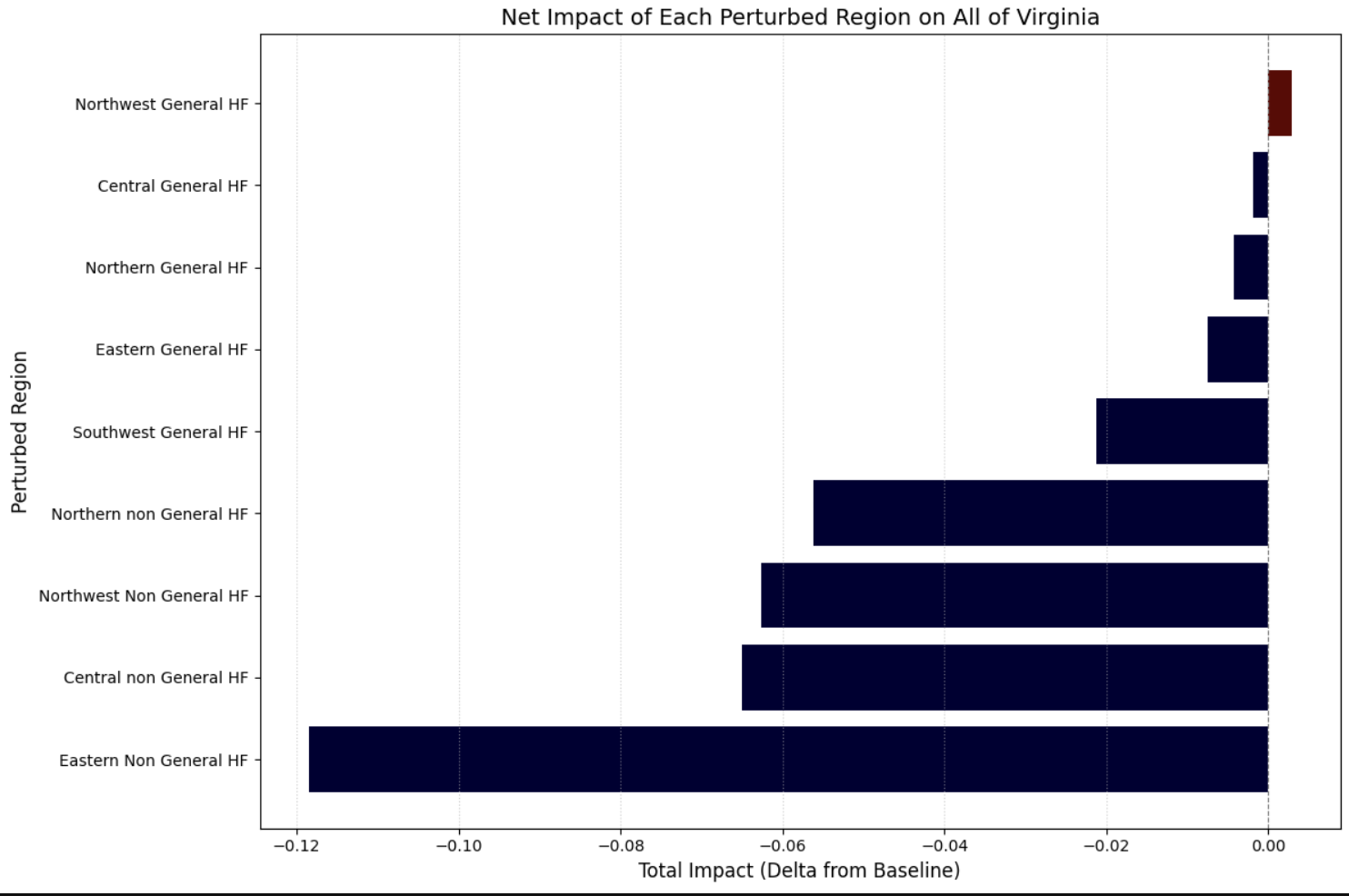}
\caption{Effect of reducing infection rate ($\beta$) in the Eastern Non-General Health Care Facility region: significant overall infection reduction, but increased cases observed in Northwest General Health Care Facilities due to mobility-driven sustained transmission.}
\label{fig:beta_reduction_effects}
\end{figure}

\begin{figure}[h]
    \centering
    \includegraphics[width=\linewidth]{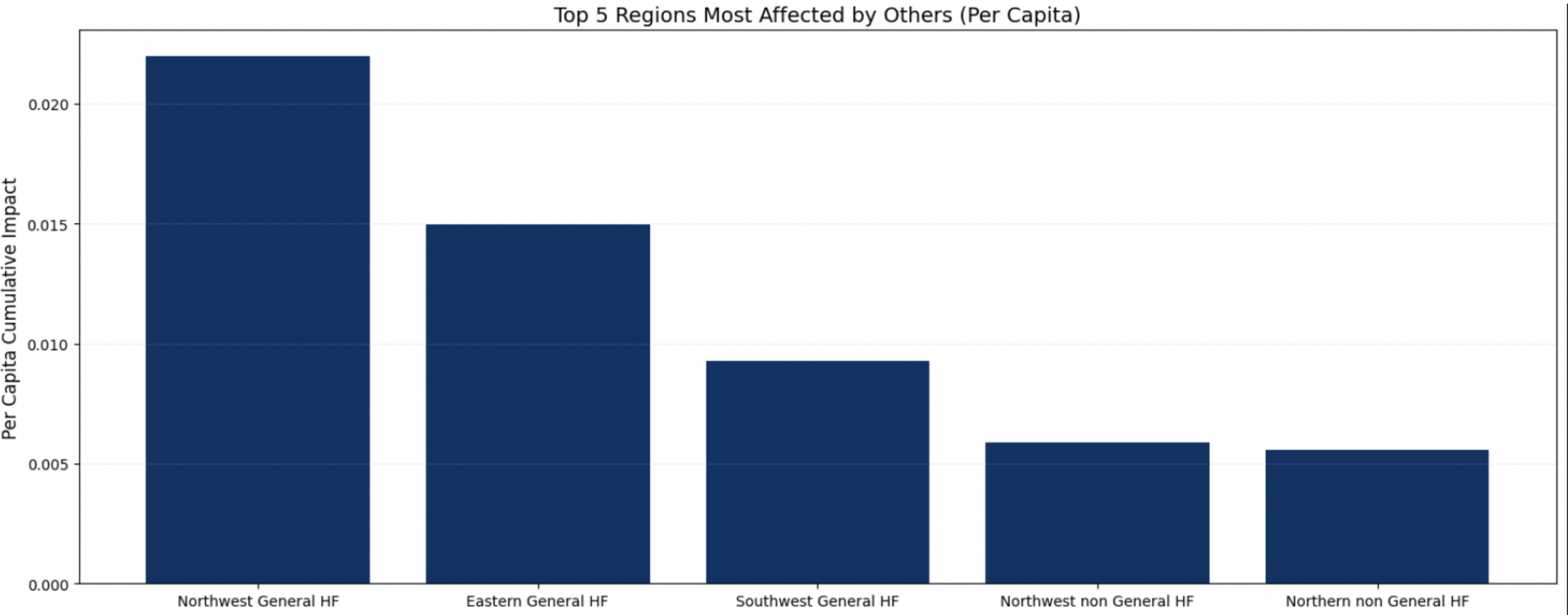}
    \caption{Top 5 regions most sensitive per capita to increased transmission elsewhere, shown by normalized infections.}
    \label{fig:sensitivity}
\end{figure}

\begin{figure}
    \centering
    \includegraphics[width=\linewidth]{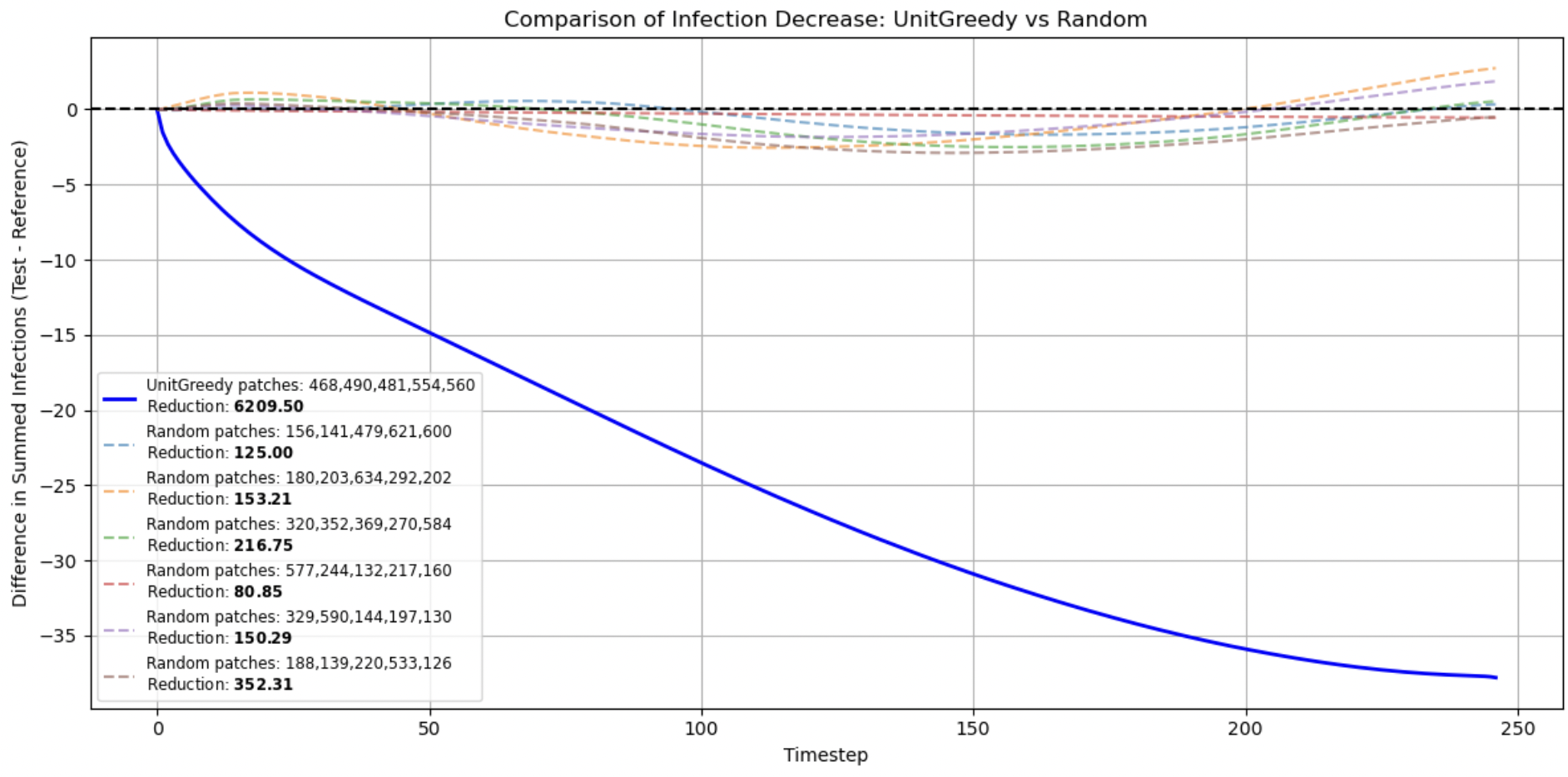}
    \caption{Comparison of infection reduction over time between the UnitGreedy-selected patches (solid blue line) and random patch selections (dashed lines)}
    \label{fig:unitgreedy}
\end{figure}

\subsection{Epidemic analyses using learned $\textbf{M(.)}$}
\paragraph{Optimizing Infection Control Policy}
Figure \ref{fig:budget} illustrates the infection reduction achieved under different intervention budgets, where the budget represents the number of healthcare patches selected by the UnitGreedy algorithm. Each successive increase in budget produces a substantial gain: targeting a single patch reduces infections by about 2,285, while expanding to five patches yields a cumulative reduction of over 6,190. This corresponds to nearly a $2.7\times$ improvement in outbreak reduction as the budget increases from 1 to 5. Importantly, the gains follow a near-linear trend, showing that the greedy algorithm consistently identifies patches with strong marginal contributions rather than exhausting high-value locations too quickly. This behavior highlights the ability of UnitGreedy to prioritize high-risk patches efficiently, offering a scalable alternative to brute-force search, $160\times$ fewer evaluations: $\binom{644}{2} \approx 207{,}000$ vs.\ 1,287; brute-force $\mathcal{O}(N^K)$ vs.\ greedy $\mathcal{O}(N K)$.

In addition, Figure \ref{fig:unitgreedy} compares the infection reduction trajectories of UnitGreedy-selected patches (solid blue line) against random patch selections (dashed lines). The greedy approach consistently outperforms random allocation, achieving faster and larger reductions over time, further demonstrating its effectiveness for targeted intervention planning.

\begin{figure}
    \centering
    \includegraphics[width =1\linewidth]{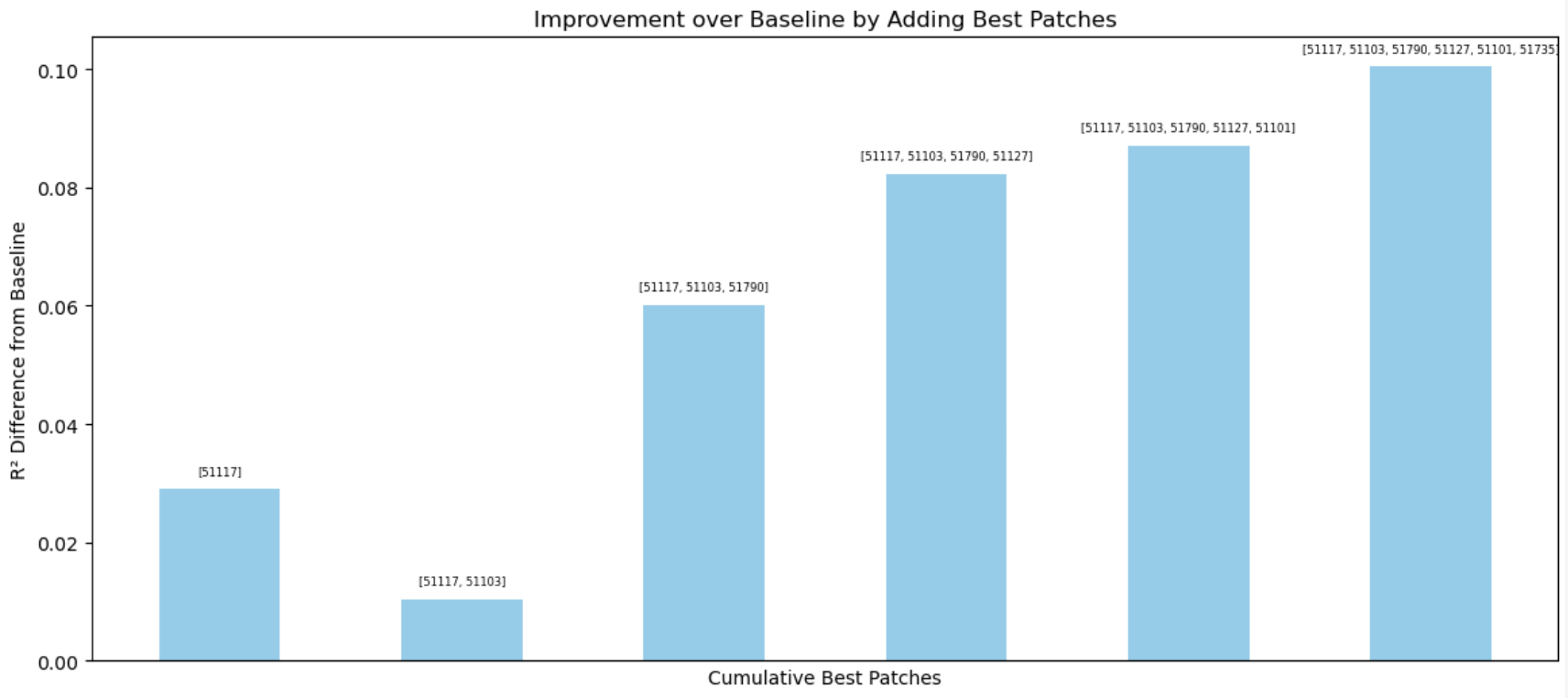}
    \caption{Improvement in $R^2$ over baseline as HF (healthcare facilities) patches are sequentially corrected. Bars show cumulative $R^2$ gain when noisy patches are replaced in greedy order; baseline $R^2$ corresponds to all HF patches being noisy during training.}
    \label{fig:HF_correct}
\end{figure}

\paragraph{Effect of Correcting HF (health care Facility) Data}
Figure \ref{fig:HF_correct}, demonstrates the impact of data quality on model performance. Starting from a baseline where all healthcare facility patches were corrupted with Gaussian noise (std =$ 0.2, \approx 3\%$ of the average infection count) during training (while normal counties remained correct), the model achieved an $R^2$ of $0.62$ when evaluated on the true, clean data. We then applied a greedy strategy: sequentially replacing the noisy HF patches with their correct data, one patch at a time, selecting the patch that yields the largest improvement at each step. The plot shows a clear cumulative improvement in $R^2$, indicating that some HF patches have a disproportionately large effect on model accuracy. This highlights that obtaining accurate data from high-impact healthcare facilities is critical for reliable forecasting. Notably, after correcting just a few key patches, the $R^2$ approaches the level achieved with all correct data, emphasizing that targeted data collection can significantly enhance predictive performance while potentially reducing overall data acquisition costs.
Only a few high-impact corrections are sufficient to restore accuracy near the fully correct baseline. This result underscores the critical role of reliable HF-level data in statewide forecasting accuracy.

\begin{figure*}
    \centering
    \includegraphics[width=1\linewidth]{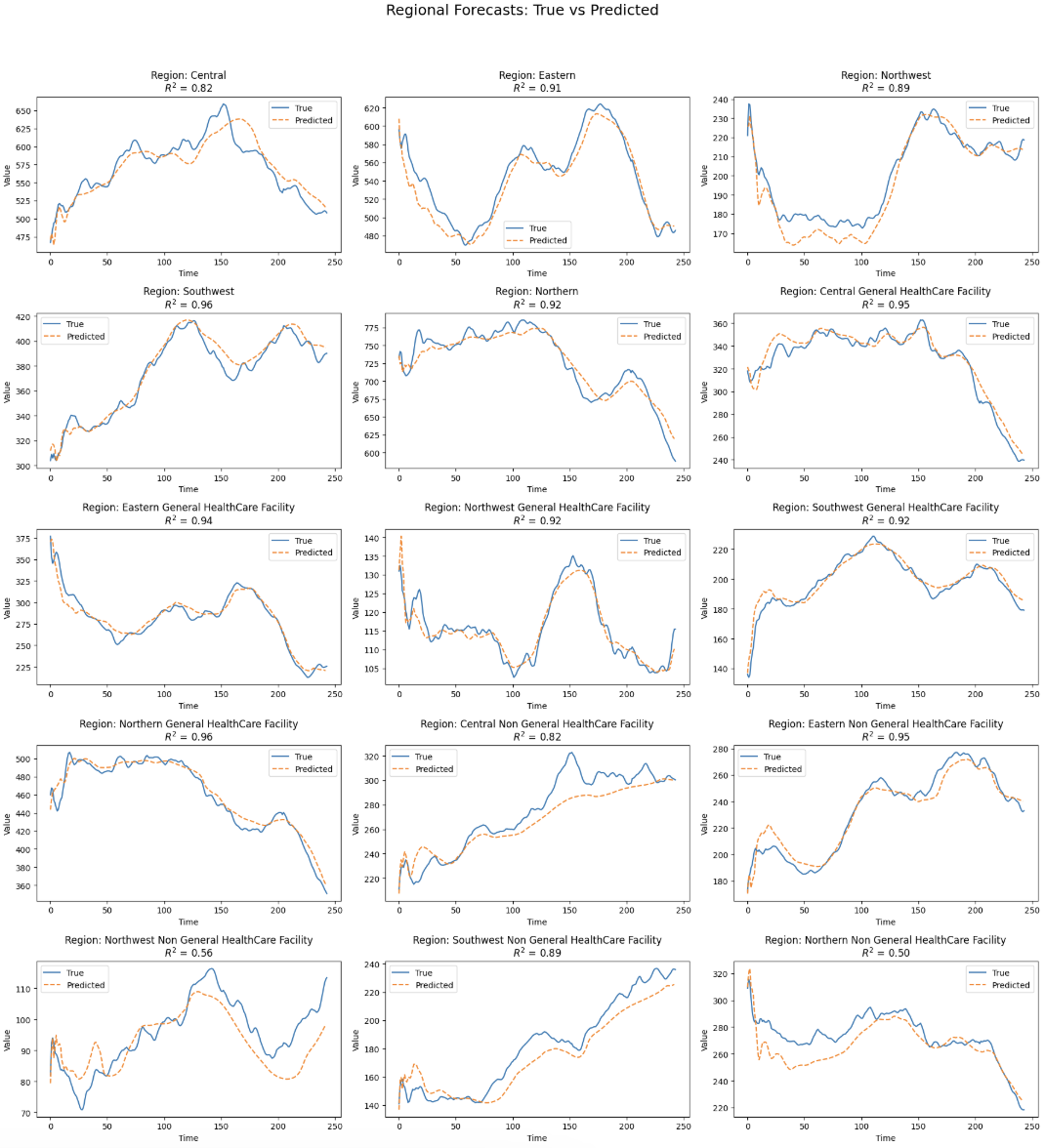}
    \caption{Comparison of observed and predicted MRSA cases at the regional level using \tool{}, $M(.)$}
    \label{fig:all_region}
\end{figure*}

\end{document}